# Chiplet Placement Order Exploration Based on Learning to Rank with Graph Representation

Zhihui Deng[1†], Yuanyuan Duan[2†], Leilai Shao[1]*, Xiaolei Zhu[2]*
[1]School of Mechanical Engineering, Shanghai Jiao Tong University, Shanghai, China
[2]School of Micro-Nano Electronics, Zhejiang University, Hangzhou, China

*Abstract*—Chiplet-based systems, integrating various silicon dies manufactured at different integrated circuit technology nodes on a carrier interposer, have garnered significant attention in recent years due to their cost-effectiveness and competitive performance. The widespread adoption of reinforcement learning as a sequential placement method has introduced a new challenge in determining the optimal placement order for each chiplet. The order in which chiplets are placed on the interposer influences the spatial resources available for earlier and later placed chiplets, making the placement results highly sensitive to the sequence of chiplet placement. To address these challenges, we propose a learning to rank approach with graph representation, building upon the reinforcement learning framework RLPlanner. This method aims to select the optimal chiplet placement order for each chiplet-based system. Experimental results demonstrate that compared to placement order obtained solely based on the descending order of the chiplet area and the number of interconnect wires between the chiplets, utilizing the placement order obtained from the learning to rank network leads to further improvements in system temperature and inter-chiplet wirelength. Specifically, applying the top-ranked placement order obtained from the learning to rank network results in a 10.05% reduction in total inter-chiplet wirelength and a 1.01% improvement in peak system temperature during the chiplet placement process.

## I. INTRODUCTION

As device dimensions approach physical limits, the sustainability of Moore's Law in driving advancements in integrated circuits becomes increasingly challenging. Chiplet-based systems, integrating diverse silicon dies manufactured at different integrated circuit process nodes onto carrier interposer layers, have gained substantial attention in recent years due to their cost-effectiveness and competitive performance [1], [2]. The core task during the chiplet placement phase is to determine the specific positions of each chiplet while achieving objectives such as minimizing the total wirelength as much as possible [3]. With the increasing complexity and compactness of chiplet-based systems, careful consideration of interconnection delay and thermal issues is crucial during the placement phase [4], [5]. These issues can be measured through total wirelength and peak temperature.

In recent years, reinforcement learning (RL) methods have emerged as promising on-chip placement optimization techniques [6-8]. Since RL iteratively completes the placement by placing one chiplet at a time, the sequence of chiplet placement becomes a significant factor influencing the final placement results. A common practice is to place larger chiplets first, as this allows for more available positions to accommodate larger chiplets without overlap [6], [8-10]. Additionally, considering the connectivity between chiplets when determining the placement order is crucial, as a higher number of interconnections reflects stronger dependencies between chiplets [6], [10]. By calculating the weighted sum of the chiplet area and the number of interconnections between chiplets, followed by descending order arrangement, it is easy to determine the placement order for each chiplet. However, selecting different weighting coefficients may lead to vastly different placement orders and outcomes.

For each chiplet-based system[5, 10], RLPlanner [10] can yield results under any placement order, and these results can be visualized as a scatter plot, as shown in Fig.1. Each point on the plot corresponds to a specific placement order. The scatter points in the lower-left corner form a Pareto curve, achieving a trade-off between the system's peak temperature and the total wirelength. The red pentagon represents the results corresponding to the placement order calculated by RLPlanner, considering the importance score $PR(v_i)$ of interconnections between chiplets multiplied by the area $Area(v_i)$ of chiplet $v_i$ [10]. It can be observed that there is a certain distance from the Pareto frontier, indicating room for further optimization.

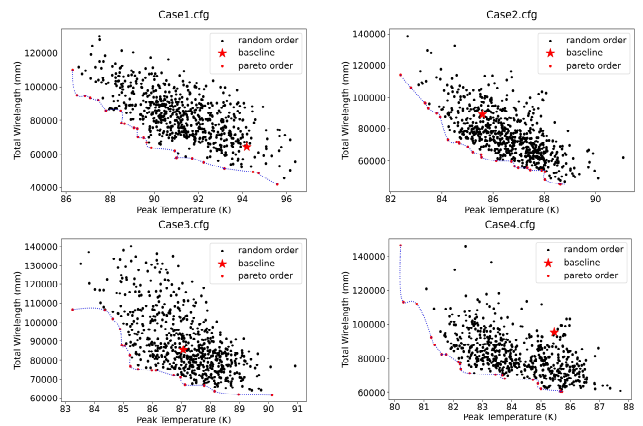

Fig. 1. Visualization of placement solution with random ordering for four typical chiplet-based systems in preivous literature.

The quantitative analysis of chiplet placement order sensitivity is illustrated in Fig. 2. For the four chiplet-based systems depicted in the figure, the histogram distribution of temperature and wirelength exhibits a trend of higher frequency in the middle and lower frequency on the sides, indicating that the results are relatively consistent for most chiplet placement orders. It becomes evident that only through more efficient and high-quality ordering strategies can further optimization of placement results be achieved. Hence, exploring an efficient and rapid method to identify the optimal placement order corresponding to Pareto frontier points for each chiplet-based system is a crucial research problem.

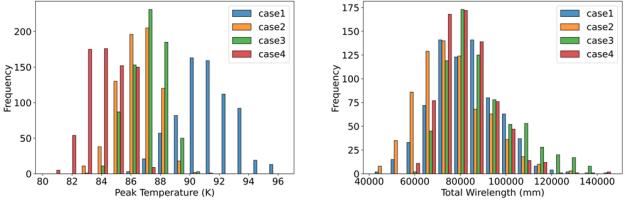

Fig. 2. Histograms of frequency of peak temperature (left) and total wirelength (right).

Existing literature primarily focuses on RL-based placement methods when it comes to placement order issues [6-8]. In the context of the placement order, these methods typically arrange modules in ascending or descending order based on a single criterion, encompassing factors like module size, module topological order, and interconnections count. DREAMPlace [11] is an advanced placement method, where the placement order of standard cells is determined solely by their area, with larger cells given priority. Google's Mirhoseini et al. [6] proposed an RL method for macro placement, considering both topological order and area when determining the macro placement order. Cheng et al. [7] introduced two joint learning methods, DeepPlace and DeepPR, where the former integrates RL and gradient-based optimization to placement macros and standard cells, and the latter employs RL for macro placement and routing. While both methods inevitably involve placement order issues for macros and standard cells, the authors did not explicitly mention it in the papers. Lai et al. [8] introduced the RL-based MaskPlace algorithm, transforming macro placement into a problem of pixel-level visual representation learning for circuit modules. This approach considers net count, module area, and the placement status of connected modules when determining placement order. Shi et al. [9] proposed a new Black-Box Optimization (BBO) framework for macro placement tasks, known as WireMask-BBO. This placement framework objectively evaluates each step's wirelength increment using a greedy algorithm guided by wire masks, inspired by Lai et al. [8]. In this work, the placement order for macros is determined by calculating the total area of standard cells connected to each macro and arranging them in descending order.

In this paper, we propose a chiplet placement order ranking framework that utilizes a graph neural network to simultaneously capture chiplet size, power consumption, and interconnect information and can efficiently learn the relative merits of chiplet placement order through the Learning to Rank (LTR) method [12]. Ultimately, the framework can select the optimal set of chiplet placement orders for each chiplet-based system, achieving a trade-off between total wirelength and peak system temperature. This approach provides robust support for early-stage chiplet placement decisions, with the potential to enhance efficiency and performance throughout the entire chiplet-based system design process. The main contributions of this paper can be summarized as follows:

- We propose a learning to rank method with graph representation capable of selecting the optimal set of chiplet placement orders for each chiplet-based system. To the best of our knowledge, this is the first dedicated work specifically studying placement orders. We developed an algorithm that assigns relevance to all placement orders corresponding to scatter points based on their distance from the Pareto frontier. Additionally, we created a large dataset to train and test our learning to rank method with graph representation.

- We designed a novel network architecture with RankNet as the main network, incorporating a feature extractor composed of multiple layers of GraphSage, a mean pooling layer, and fully connected layers. In this work, chiplet-based systems with connectivity relationships are treated as undirected graphs, and graph neural networks are utilized for feature learning, with the obtained graph-level representations serving as input to RankNet.

- Regarding feature selection, we considered not only the dimensions, power, and interconnection count between chiplets, as in previous works, but also incorporated changes in on-chip available area, total power, and interconnection wirelength. The latter provides the model with a forward-looking perspective, allowing it to anticipate changes in the placement results of chiplet-based systems.

- Experimental results demonstrate that compared to placement orders obtained by considering only chiplet area and interconnection count in descending order, using the placement orders obtained from the LTR network in the chiplet placement process further improves system temperature and interconnection wirelength. Specifically, applying the top-ranked placement order obtained from the learning to rank network results in a 10.05% improvement in total wirelength and a 1.01% improvement in peak system temperature.

The remainder of this paper is organized as follows: Section II introduces the prerequisites of RLPlanner. Section III provides a detailed description of our learning to rank method with graph representation. Section IV presents the validation of our model in chiplet placement tasks. Section V concludes the paper.

II. PRELIMINARIES

In this section, we introduce the background of RLPlanner. As shown in Fig.3, the chiplet placement and routing process based on RL is illustrated [10]. The overall architecture of the RLPlanner consists of three main components: the chiplet placement environment, an RL-based agent, and a thermally aware reward calculator. The placement environment updates action mask $M_{t+1}$ and state $s_t$ for the already placed chiplets at each time step $t$. The agent takes the previous state $s_{t-1}$ and action mask $M_t$ as input and generates action probability matrices $\pi_\theta(a_t|s_t)$ and expected rewards $V_\phi(s_t)$ using policy and value networks, respectively. The probabilities of infeasible actions are set to "0" based on the action mask before sampling actions. Subsequently, action $a_t$ are sampled from the updated probability matrices $\tilde{\pi}_\theta(a_t|s_t)$. Once all chiplets are placed, the reward calculator initiates microbump allocation, minimizing the total wirelength by assigning positions for the pins connected between each pair of chiplets. Following microbump allocation, temperature prediction is performed.

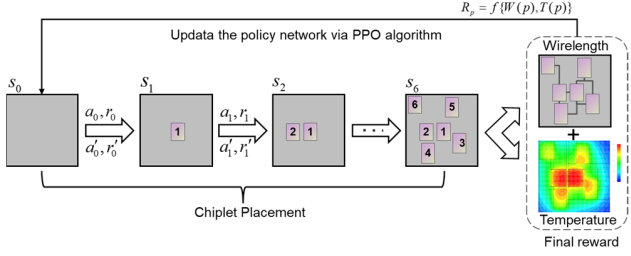

Fig. 3. Chiplet placement and routing process based on reinforcement learning.

## III. METHODOLOGY

Fig.4 illustrates the chiplet placement order exploration process based on the LTR network with graph representation. The process includes generating training and testing datasets, partitioning the training dataset into strongly correlated and weakly correlated sets, constructing a graph data structure, model training, and model testing. The following provides a detailed explanation.

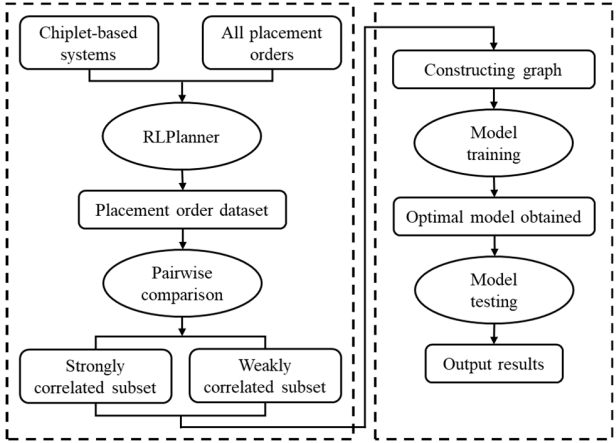

Fig. 4. Flow chart of chiplet placement order exploration based on learning to rank with graph representation.

### A. Generating Datasets

Based on RLPlanner, we construct chiplet placement order datasets for different chiplet-based systems under all possible permutations of placement orders. We extract the total wirelength and peak temperature from the placement results generated by RLPlanner. All cases are derived from [5] and [10], partly open-source chiplet-based systems and partly synthesized. Cases 1-4 are the training dataset and the others are the testing dataset. Using these two metrics, we determine the correlation and categorize the training dataset into strongly correlated and weakly correlated subsets through pairwise comparisons. For example, for a design with 6 chiplets, there are a total of 720 different placement orders. Running RLPlanner with these placement orders yields various placement results, which can be visualized in a scatter plot. Assigning a correlation value to each scatter point based on its distance from the Pareto frontier, where closer points have higher correlation, we define correlation levels from 1 to 10 (with 10 being the highest). The distribution of points in each correlation level is summarized in Table 1.

TABLE I. STATISTICS ON THE NUMBER OF SCATTERS AT EACH DEGREE OF CORRELATION

| Correlation $L$ | $\Delta$ | Case1 | Case2 | Case3 | Case4 |
|---|---|---|---|---|---|
| 1 | 0.1 | 53 | 86 | 97 | 74 |
| 2 | 0.2 | 45 | 85 | 88 | 92 |
| 3 | 0.3 | 75 | 118 | 102 | 128 |
| 4 | 0.4 | 109 | 117 | 109 | 108 |
| 5 | 0.5 | 119 | 101 | 78 | 90 |
| 6 | 0.6 | 98 | 79 | 81 | 90 |
| 7 | 0.7 | 72 | 49 | 58 | 65 |
| 8 | 0.8 | 56 | 34 | 37 | 31 |
| 9 | 0.9 | 44 | 22 | 31 | 15 |
| 10 | other | 50 | 30 | 40 | 28 |

The correlation assignment rule for scatter points is as follows: different values $\Delta$ from Table I are taken, and the correlation $L$ corresponding to the scatter point $(T_i, WL_i)$ meeting Equation (1) is given by the first column. When $\Delta = 0$, Equation (1) corresponds to the definition of the Pareto curve, indicating that there are no other points simultaneously better than the point represented by the Pareto curve in both temperature and wirelength metrics.

$$T_i \leq T_j + \Delta_1 \quad or \quad WL_i \leq WL_j + \Delta_2, \quad \exists j \in 1,2,\ldots,N \quad (1)$$

$$\Delta_1 = \Delta(\frac{1}{N_2}\sum_{l \in Q}T_l - \frac{1}{N_1}\sum_{k \in P}T_k), \quad \Delta_2 = \Delta(\frac{1}{N_2}\sum_{l \in Q}WL_l - \frac{1}{N_1}\sum_{k \in P}WL_k) \quad (2)$$

Where $P$ and $Q$ are the sets formed by scatter points in the bottom-left and top-right corners, respectively. $N_1$ is the number of points in set $P$, $N_2$ is the number of points in set $Q$, and $\Delta_1$ and $\Delta_2$ are the deviations calculated based on temperature $T$ and wirelength $WL$, respectively.

For each pair of result points within the same chiplet-based system, they are compared against each other, and based on the correlation, they are categorized into strongly correlated and weakly correlated sets, forming a training data point together. To limit the sampling scope and ensure the dataset's manageable size, each resulting point within the same chiplet-based system can only be compared with up to $k$ (where $k = 10$) other points.

### B. Constructing Graph

Fig.5 illustrates the chiplet-based system can be abstracted into an undirected graph based on the following rules:

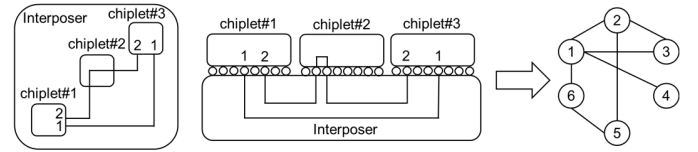

Fig. 5. Construction graph from the chiplet-based system.

each chiplet is defined as a node of the same type, and the interconnection relationships determine the edges between nodes. As the chiplet interconnections are bidirectional, the resulting graph is undirected. Fig.5 left and middle depict the top view and cross-sectional view of the chiplet-based system, respectively, with only three chiplets shown for clarity. Fig.5 right represents the entire chiplet-based system abstracted into the graph.

*C. Constructing Node Feature Vectors*

As shown in Table II, each node in the constructed graph is characterized by a 6-dimensional vector, including chiplet width, length, power, corresponding placement order, changes in the available area, total power, and interconnect wirelength associated with placing the chiplet in that order. The edge feature is the number of interconnect wires, and the graph label is the correlation calculated based on the distance to the Pareto front when generating the training dataset. The "Corresponding placement order" indicates the step at which the chiplet is placed. "Available area change" represents the change in the remaining placement space after placing the chiplet. "Total power change" represents the change in the total power of placed chiplets. "Interconnect wirelength change" represents the change in interconnect wirelength between placed chiplets. For ease of model training, both node and edge features of the graph undergo min-max normalization.

TABLE II. FEATURES OF EACH GRAPH

| Type | Features |
|---|---|
| node feature | width (mm) |
| | length (mm) |
| | power (W) |
| | corresponding placement order |
| | available area change ($mm^2$) |
| | total power change (W) |
| | interconnect wirelength change (mm) |
| edge feature | interconnect count |
| graph label | correlation |

*D. Model Training*

Each chiplet-based system's circuit, combined with a specific placement order, can be abstracted into a unique graph. The model is trained using a LTR network with graph representation, as shown in Fig.6.

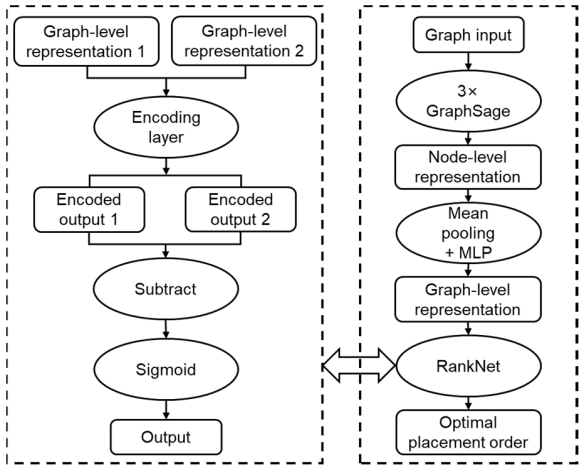

Fig. 6. Learning to rank network with graph representation.

The model architecture is based on the RankNet [12] backbone, with a feature extractor composed of graph convolutional layers, a mean pooling layer, and fully connected layers to obtain graph-level representations as input for RankNet. The graph convolutional layers consist of a stack of three GraphSage [13] layers with neuron quantities of 7, 32, and 64, respectively. These graph convolutional layers are employed to learn node representations of the graph, and the graph-level representation is obtained through mean pooling. The obtained graph-level representation undergoes three layers of fully connected layers forming a Multilayer Perceptron (MLP) with neuron quantities of 64, 32, and 16. ReLU activation functions are applied between layers, the batch size for training sets is set to 64 and the loss function is binary cross-entropy. The total number of iterations is set to 3000, and the Adam [14] optimizer with a learning rate of 0.0001 is employed.

IV. EXPERIMENTS AND RESULTS

Our model was implemented using the PyTorch [15] machine learning framework on a Linux workstation equipped with an Intel Xeon Gold 6230 (2.1GHz) CPU, 128GB of RAM, and an NVIDIA Quadro RTX5000 graphic card.

*A. Comparison between the two ordering methods*

TABLE III. THE RESULTS COMPARISON BETWEEN THE TWO ORDERING METHODS

| | Baseline | | | Learning to Rank (top-ranked) | | |
|---|---|---|---|---|---|---|
| | T(°C) | WL(mm) | L | T(°C) | WL(mm) | L |
| Case 1 | 94.19 | 64333.75 | 6 | 89.02 | 86754.86 | 8 |
| Case 2 | 85.58 | 88909.26 | 3 | 86.17 | 61504.74 | 9 |
| Case 3 | 87.06 | 85404.31 | 6 | 86.36 | 77996.53 | 8 |
| Case 4 | 85.45 | 95573.08 | 0 | 83.86 | 69854.69 | 9 |
| Training-average | 88.07 | 83555.10 | 3.7 | **86.35** (-1.95%) | **74027.70** (-11.40%) | 8.5 |
| Case 5 | 86.19 | 88923.56 | 7 | 86.30 | 80616.43 | 8 |
| Case 6 | 87.13 | 59957.18 | 7 | 86.66 | 52971.71 | 8 |
| Case 7 | 83.86 | 78756.66 | 4 | 81.64 | 71162.38 | 9 |
| Testing-average | 85.73 | 75879.13 | 6 | **84.86** (-1.01%) | **68250.17** (-10.05%) | 8.3 |

* "T(°C)" represents the peak system temperature. "WL(mm)" represents the total wirelength. "L" represents the correlation.

We validate the trained model on the test dataset, using RLPlanner with the placement order obtained from descending $PR(v_i) \times Area(v_i)$ as a baseline. By employing two different methods to calculate chiplet placement orders and applying them to RLPlanner, a comparison is made based on the total wirelength and peak temperature, as shown in Table III. It can be observed that the placement results obtained by applying the placement order derived from the LTR network are closer to the Pareto front, with better-combined performance for both metrics. Notably, cases 3-7 show improvements in both peak temperature and interconnect wirelength compared to the baseline method. For case 1, a reduction of 5°C in peak temperature is achieved at the cost of increasing the wirelength by 22400mm. This trade-off is due to the high power consumption of this chiplet-based system, requiring the chiplets to be spread out as much as possible to maintain thermal stability. Conversely, case 2 achieves a decrease of 27400mm in interconnect wirelength at the expense of a 0.5°C increase in temperature. This trade-off is easily understood, as a slight increase in temperature within the threshold range does not affect normal operation while reducing wirelength can improve the performance of the chiplet-based system.

Overall, applying the top-ranked placement order obtained from the LTR network results in a 1.01% improvement in peak temperature and a 10.05% improvement in interconnect wirelength. This highlights the effectiveness of the algorithm proposed in this paper.

TABLE IV. FURTHER COMPARISON BETWEEN THE TWO ORDERING METHODS

|  | Baseline | | | Learning to Rank (top 5 ranked) | | |
|---|---|---|---|---|---|---|
|  | T(℃) | WL(mm) | L | T(℃) | WL(mm) | L |
| Case 1 | 94.19 | 64333.75 | 6 | 87.8 | 96967.94 | 8.4 |
| Case 2 | 85.58 | 88909.26 | 3 | 86.94 | 59747.97 | 8.6 |
| Case 3 | 87.06 | 85404.31 | 6 | 87.26 | 74792.84 | 8.0 |
| Case 4 | 85.45 | 95573.08 | 0 | 84.06 | 70570.19 | 8.8 |
| Training-average | 88.07 | 83555.10 | 3.7 | **86.51** (-1.77%) | **75519.73** (-9.62%) | 8.5 |
| Case 5 | 86.19 | 88923.56 | 7 | 85.81 | 80904.81 | 8.6 |
| Case 6 | 87.13 | 59957.18 | 7 | 88.36 | 51949.06 | 7.0 |
| Case 7 | 83.86 | 78756.66 | 4 | 82.73 | 66979.31 | 7.6 |
| Testing-average | 85.73 | 75879.13 | 6 | **85.63** (-0.12%) | **66611.06** (-12.21%) | 7.7 |

* "T(℃)" represents the peak system temperature. "WL(mm)" represents the total wirelength. "L" represents the correlation.

Table IV provides a further comparison of the two ordering methods. The table lists the average results for the top 5 rankings obtained by the learning to rank network. Fig.7 displays the corresponding positions in the scatter plot for the top 5 predicted results of four chiplet-based systems. In summary, applying the top 5 ranked placement orders obtained from the learning to rank network results in a 0.12% improvement in peak temperature and a 12.21% improvement in interconnect wirelength.

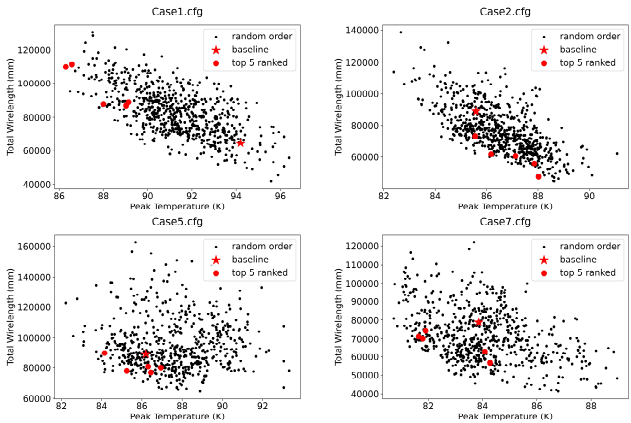

Fig. 7. Visualization of the results with the top 5 ranked placement orders for the four chiplet-based systems. Cases 1 and 2 are the training dataset and Cases 5 and 7 are the testing dataset.

Fig.8 illustrates the results obtained for chiplet-based system 1 using different placement orders. In (a), the placement order corresponds to the baseline. In (b)-(f), the placement orders correspond to the top 5 obtained through the learning to rank network. It can be observed that (b) has the highest peak temperature among these six placement results, sacrificing thermal stability to minimize interconnect wirelength. Although (e) and (f) have different placement orders leading to varying placement results, they achieve the same peak temperature and interconnect wirelength (T = 87.46°C, WL = 92317.86mm), with the lowest temperature in this case.

Overall, applying placement orders obtained from the learning to rank network in the reinforcement learning framework can generate a set of placement results close to the Pareto front. These results achieve different trade-offs between wirelength and temperature. Chiplet-based system designers can utilize this visual representation to explore and select placement results that best align with their specific design constraints and preferences.

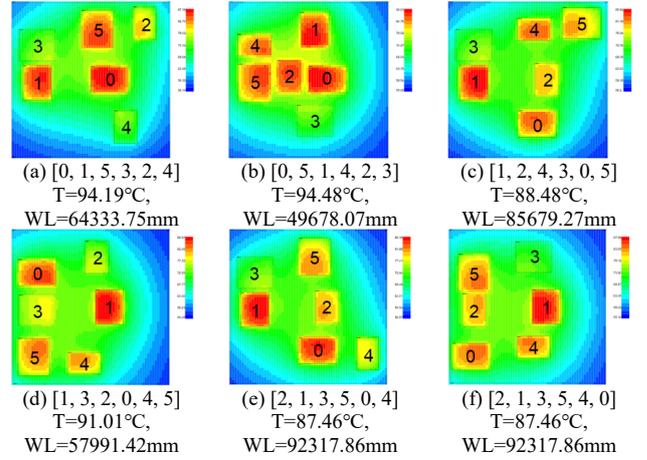

(a) [0, 1, 5, 3, 2, 4] T=94.19°C, WL=64333.75mm
(b) [0, 5, 1, 4, 2, 3] T=94.48°C, WL=49678.07mm
(c) [1, 2, 4, 3, 0, 5] T=88.48°C, WL=85679.27mm
(d) [1, 3, 2, 0, 4, 5] T=91.01°C, WL=57991.42mm
(e) [2, 1, 3, 5, 0, 4] T=87.46°C, WL=92317.86mm
(f) [2, 1, 3, 5, 4, 0] T=87.46°C, WL=92317.86mm

Fig. 8. The placement results of case 1 were obtained using different placement orders. (a) represents baseline placement order. (b)-(f) represents the top 5 ranked placement orders obtained through the learning to rank network.

The framework for chiplet placement order exploration based on the learning to rank network can achieve success, and the selection of features and the design of the network architecture are inherently intertwined. Regarding feature selection, we incorporated changes in on-chip available area, total power, and interconnection wirelength. These features provide the model with a forward-looking perspective, enabling it to anticipate changes in the placement results of chiplet-based systems. In terms of architecture design, the machine learning model employed in this paper combines graph convolutional networks and learning to rank networks. This architectural design maximizes the advantages of both components.

*B. Ablation study*

Table V illustrates the application of different pooling layers to obtain graph-level representations in the learning to rank network. The obtained placement orders from the top-ranked are then compared in RLPlanner for peak temperature and interconnect wirelength. The results indicate that placement orders obtained with three different pooling layers can all generate placement results close to the Pareto front. The effectiveness of the learning to rank network is consistent across the different pooling layers.

TABLE V. THE RESULTS COMPARISON USING DIFFERENT POOLING METHODS

|  | Sum Pooling | | Max Pooling | | Mean Pooling | |
| --- | --- | --- | --- | --- | --- | --- |
|  | T(℃) | WL(mm) | T(℃) | WL(mm) | T(℃) | WL(mm) |
| Case 1 | 89.65 | 66973.81 | 88.50 | 78249.47 | 89.02 | 86754.86 |
| Case 2 | 83.33 | 96548.45 | 83.46 | 93438.03 | 86.17 | 61504.74 |
| Case 3 | 90.11 | 61873.84 | 88.14 | 63921.72 | 86.36 | 77996.53 |
| Case 4 | 82.23 | 76853.27 | 81.60 | 82353.99 | 83.86 | 69854.69 |
| Case 5 | 86.84 | 67107.72 | 85.48 | 75744.27 | 86.30 | 80616.43 |
| Case 6 | 87.98 | 44385.82 | 85.53 | 53698.14 | 86.66 | 52971.71 |
| Case 7 | 81.98 | 63260.54 | 82.46 | 60470.90 | 81.64 | 71162.38 |
| Average | 86.02 | 68143.35 | 85.02 | 72553.79 | 85.72 | 71551.62 |

* "T(℃)" represents the peak system temperature. "WL(mm)" represents the total wirelength.

Table VI presents a comparison of the results of the learning to rank network trained with different sampling sizes during the process of pairwise comparisons between every two result points within the same chiplet-based system. For $k = 1$, only adjacent result points are compared, while $k = 10, 20$ represent sampling 10, 20 result points for comparison, respectively. In the extreme case, each resulting point is compared with all other objects, resulting in $C_{720}^2$ training data points for a system with 6 chiplets. Such a scenario leads to a dataset with significant redundancy and data explosion, considerably prolonging the training time of the learning to rank network. It is observed that $k = 10$ strikes a balance between training time and model performance. Therefore, in this work, $k = 10$ was chosen to limit the sampling scope, ensuring that the dataset remains manageable without compromising model effectiveness.

TABLE VI. THE RESULTS COMPARISON USING DIFFERENT SAMPLE SIZE $k$

|  | $k = 1$ | | $k = 10$ | | $k = 20$ | |
| --- | --- | --- | --- | --- | --- | --- |
|  | T(℃) | WL(mm) | T(℃) | WL(mm) | T(℃) | WL(mm) |
| Case 1 | 91.14 | 87773.37 | 89.02 | 86754.86 | 89.91 | 90978.35 |
| Case 2 | 86.36 | 81498.94 | 86.17 | 61504.74 | 84.23 | 89198.47 |
| Case 3 | 87.91 | 79024.02 | 86.36 | 77996.53 | 88.28 | 83802.75 |
| Case 4 | 85.14 | 63010.21 | 83.86 | 69854.69 | 82.05 | 86956.82 |
| Case 5 | 86.28 | 72246.60 | 86.30 | 80616.43 | 88.09 | 73839.13 |
| Case 6 | 87.65 | 63359.35 | 86.66 | 52971.71 | 90.52 | 34694.12 |
| Case 7 | 85.41 | 65567.45 | 81.64 | 71162.38 | 82.01 | 64330.73 |
| Average | 87.12 | 73211.42 | 85.72 | 71551.62 | 86.44 | 74828.62 |

* "T(℃)" represents the peak system temperature. "WL(mm)" represents the total wirelength.

Simultaneously, the process of obtaining chiplet placement orders through learning to rank can be completed within one second, rendering the introduced time overhead negligible when compared to the actual placement process. While obtaining all placement results for a chiplet-based system requires running through all possible placement orders. collecting results for all chiplet-based systems, and subsequently compiling training data, this process takes a comparatively longer time. Fortunately, since the model operates offline after training, the data acquisition task needs to be performed only once before the formal training of the model. In summary, this paper provides researchers employing reinforcement learning for sequential placement with a viable solution for determining placement order.

## V. CONCLUSIONS

In this paper, we propose a learning to rank method with graph representation capable of selecting the optimal set of chiplet placement orders for each chiplet-based system. We designed a novel network architecture with RankNet as the backbone, incorporating a feature extractor composed of graph convolutional layers, a mean pooling layer, and fully connected layers to obtain graph-level representations. Experimental results demonstrate that utilizing the placement order obtained from the learning to rank network leads to further improvements in system temperature and inter-chiplet wirelength. In the future, we will improve the method in this paper to cope with macro or standard cell placement tasks.

ACKNOWLEDGMENT

This work is supported by the National Key R&D Program of China (2023YFB4402700), the Major Scientific Research Project of Zhejiang Province (No.2022C01048) and the Pre-research project (No.31513010501).

REFERENCES

[1] R. Farjadrad, M. Kuemerle and B. Vinnakota, "A Bunch-of-Wires (BoW) Interface for Interchiplet Communication," in IEEE Micro, vol. 40, no. 1, pp. 15-24, 1 Jan.-Feb. 2020.
[2] A. Kannan et al., "Enabling interposer-based disintegration of multicore processors," in Proceedings of the 48th international symposium on Microarchitecture, 2015, pp. 546–558.
[3] R. Radojcic, More-than-Moore 2.5 D and 3D SiP Integration. Springer, 2017.
[4] F. Eris et al., "Leveraging thermally-aware chiplet organization in 2.5 d systems to reclaim dark silicon," in 2018 Design, Automation & Test in Europe Conference & Exhibition (DATE). IEEE, 2018, pp. 1441–1446.
[5] Y. Ma et al., "TAP-2.5D: A Thermally-Aware Chiplet Placement Methodology for 2.5D Systems," 2021 Design, Automation & Test in Europe Conference & Exhibition (DATE), Grenoble, France, 2021, pp. 1246-1251.
[6] A. Mirhoseini et al., "A graph placement methodology for fast chip design," Nature, vol. 594, no. 7862, pp. 207–212, 2021.
[7] R. Cheng et al., "On joint learning for solving placement and routing in chip design," Advances in Neural Information Processing Systems, vol. 34, pp. 16 508–16 519, 2021.
[8] Y. Lai, Y. Mu and P. Luo, "Maskplace: Fast chip placement via reinforced visual representation learning," Advances in Neural Information Processing Systems, vol. 35, pp. 24019-24030, 2022.
[9] Y. Shi et al., "Macro Placement by Wire-Mask-Guided Black-Box Optimization," arXiv preprint arXiv:2306.16844, 2023.
[10] Y. Duan et al., "RLPlanner: Reinforcement Learning based Floorplanning for Chiplets with Fast Thermal Analysis," arxiv preprint arxiv:2312.16895, 2023.
[11] Y. Lin et al., "DREAMPlace: Deep Learning Toolkit-Enabled GPU Acceleration for Modern VLSI Placement," 2019 56th ACM/IEEE Design Automation Conference (DAC), 2019, pp. 1-6.
[12] C. Burges et al., "Learning to rank using gradient descent," in Proceedings of the 22nd international conference on Machine learning, 2005, pp. 89-96.
[13] W. Hamilton, Z. Ying and J. Leskovec, "Inductive representation learning on large graphs," Advances in neural information processing systems, 2017.
[14] D. P. Kingma and J. Ba, "Adam: A method for stochastic optimization," arXiv preprint arXiv:1412.6980, 2014.
[15] M. Fey and J. E. Lenssen, "Fast graph representation learning with PyTorch geometric," in Proc. ICLR Workshop Representation Learn. Graphs Manifolds, 2019.